\documentclass[a4paper,fleqn]{cas-dc}
\usepackage{dblfnote}
\usepackage[]{natbib}

\usepackage{times}  % DO NOT CHANGE THIS
\usepackage{helvet}  % DO NOT CHANGE THIS
\usepackage{courier}  % DO NOT CHANGE THIS
\usepackage{algorithm}
\usepackage{algorithmic}
\usepackage{xcolor}
\usepackage{amsmath}
\usepackage{subfig}
\usepackage{amssymb}
\usepackage{wrapfig}
\usepackage{booktabs}
\usepackage{multirow}
\usepackage{makecell}
\usepackage{url}

\usepackage{subfig,graphicx}
\usepackage[T1]{fontenc}
\usepackage{mathdesign}
\usepackage{lipsum}
\usepackage{titlesec}
\newcommand{\charternumbers}{\fontfamily{bch}\selectfont}
\DeclareTextFontCommand{\textcharter}{\charternumbers}

\titleformat{\section}{\normalfont\fontsize{9pt}{12pt}\bfseries}{\textcharter{\thesection.}}{0.5em}{}
\titleformat{\subsection}{\normalfont\fontsize{8pt}{12pt}\itshape}{\textcharter{\thesubsection.}}{0.3em}{}
\titleformat{\subsubsection}{\normalfont\fontsize{8.3pt}{12pt}\itshape}{\textcharter{\thesubsubsection.}}{0.4em}{}

\titleformat{name=\section,numberless}{\normalfont\fontsize{9pt}{12pt}\bfseries}{}{0em}{}
\titleformat{name=\subsection,numberless}{\normalfont\fontsize{9pt}{12pt}\bfseries}{}{0em}{}
\titleformat{name=\subsubsection,numberless}{\normalfont\fontsize{9pt}{12pt}\bfseries}{}{0em}{}
\usepackage{hyperref}

\makeatletter
\newcommand*\bigcdot{\mathpalette\bigcdot@{.5}}
\newcommand*\bigcdot@[2]{\mathbin{\vcenter{\hbox{\scalebox{#2}{$\m@th#1\bullet$}}}}}
\makeatother

\usepackage{balance}
\usepackage{color}
\hypersetup{
  colorlinks=true,
  linkcolor=cyan,
  urlcolor=cyan,
  citecolor=cyan,
  linktoc=all
}
\usepackage{fancyhdr}
\def\tsc#1{\csdef{#1}{\textsc{\lowercase{#1}}\xspace}}
\tsc{WGM}
\tsc{QE}
\tsc{EP}
\tsc{PMS}
\tsc{BEC}
\tsc{DE}
\begin{document}
%去除第一页的页眉页脚
\thispagestyle{empty}

\let\WriteBookmarks\relax
\def\floatpagepagefraction{1}
\def\textpagefraction{.001}
\let\printorcid\relax

\shorttitle{Leveraging social media news}

% Short author这个位置是作者名字的缩写：
% \shortauthors{CV Radhakrishnan et~al.}

\title[mode = title]{Boosting the Transferability of Adversarial Attacks with Global Momentum Initialization}

\author[1]{\textcolor{black}{Jiafeng Wang}}[type=editor,
    auid=000,bioid=2,
    % prefix=Sir, 
    role=,]
\fnmark[1] 
\ead{jiafengwang21@m.fudan.edu.cn}
\credit{Conceptualization of this study, Methodology, Software}

\author[2,3]{\textcolor{black}{Zhaoyu Chen}}[style=chinese]
\fnmark[1]
\ead{zhaoyuchen20@fudan.edu.cn}

\author[2,3]{\textcolor{black}{Kaixun Jiang}}
\ead{kxjiang22@m.fudan.edu.cn}

\author[2,3]{\textcolor{black}{Dingkang Yang}}
\ead{dkyang20@fudan.edu.cn}

\author[1]{\textcolor{black}{Lingyi Hong}}
\ead{lyhong22@m.fudan.edu.cn}

\author[2,3]{\textcolor{black}{Pinxue Guo}}
\ead{pxguo21@m.fudan.edu.cn}

\author[1]{\textcolor{black}{Haijing Guo}}
\ead{hjguo22@m.fudan.edu.cn}

\author[1,2,3]{\textcolor{black}{Wenqiang Zhang}}
\cormark[1]
\ead{wqzhang@fudan.edu.cn}

\address[1]{Shanghai Key Lab of Intelligent Information Processing, School of Computer Science, Fudan University, Shanghai 200433, China}
\address[2]{Shanghai Engineering Research Center of AI \& Robotics, Academy for Engineering \& Technology, Fudan University, Shanghai 200433, China}
\address[3]{Engineering Research Center of Robotics, Ministry of Education, Academy for Engineering \& Technology, Fudan University, Shanghai 200433, China}

\fntext[1]{Equal contribution.} 
\cortext[1]{Corresponding author.} 

\begin{abstract}
Deep Neural Networks (DNNs) are vulnerable to adversarial examples, which are crafted by adding human-imperceptible perturbations to the benign inputs. Simultaneously, adversarial examples exhibit transferability across models, enabling practical black-box attacks. However, existing methods are still incapable of achieving the desired transfer attack performance. In this work, focusing on gradient optimization and consistency, we analyse the gradient elimination phenomenon as well as the local momentum optimum dilemma. To tackle these challenges, we introduce  Global Momentum Initialization (GI), providing global momentum knowledge to mitigate gradient elimination. Specifically, we perform gradient pre-convergence before the attack and a global search during this stage. GI seamlessly integrates with existing transfer methods, significantly improving the success rate of transfer attacks by an average of 6.4\% under various advanced defense mechanisms compared to the state-of-the-art method. Ultimately, GI demonstrates strong transferability in both image and video attack domains. Particularly, when attacking advanced defense methods in the image domain, it achieves an average attack success rate of 95.4\%. The code is available at \url{https://github.com/Omenzychen/
Global-Momentum-Initialization}.
\end{abstract}


\maketitle

\thispagestyle{empty}
\begin{figure*}[t]
    \centering
    \includegraphics[width=13cm,height=4.2cm]{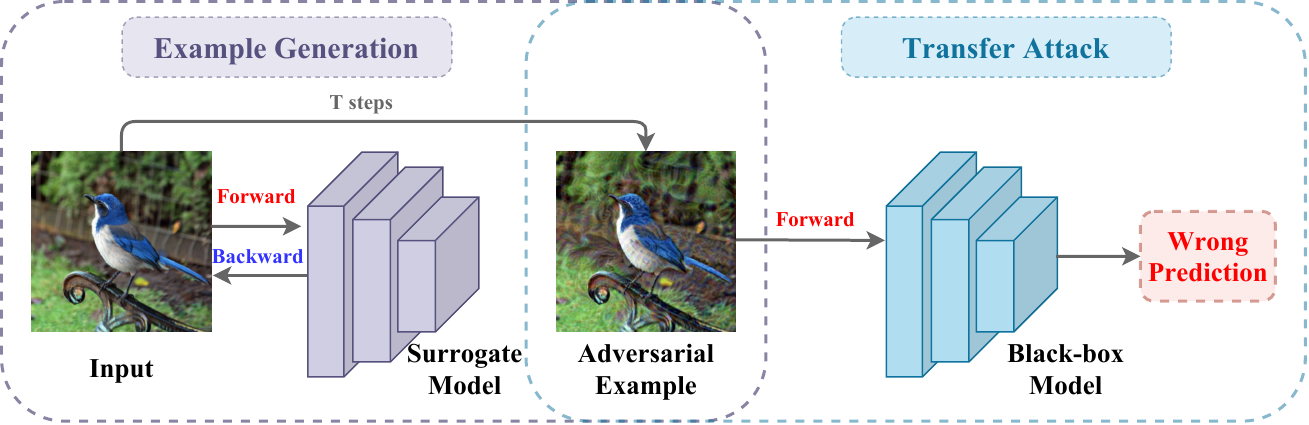}
    \caption{Pipeline of the generation and transfer process of adversarial examples. The module on the left illustrates example generation utilizing the surrogate model for $T$ iterative attacks, while the module on the right delineates the transfer of the generated adversarial examples into the black-box model.}
    \label{fig:process}
\end{figure*}

\section{Introduction}
\label{sec:intro}
Deep neural networks have shown superior performance in various tasks{~\citep{resnet,li2019learnable,bojarski2016end,wang2018cosface}, but they still exhibit high vulnerability to adversarial examples~\citep{fgsm,L_BFGS,DBLP:conf/icml/AthalyeC018,DBLP:conf/icml/LiLWZG19}, which makes the model lose its original performance by adding some imperceptible perturbations. Moreover, adversarial examples exhibit transferability across models~\citep{ Wu_Tan_Wang_Ma_Ma_Li_2024, DBLP:conf/ccs/PapernotMGJCS17,Huang2022TSEATS}, meaning adversarial examples crafted to fool one model can deceive other models, making it feasible to perform practical black-box attacks~\citep{kurakin2018adversarial}, and the attack process is shown in Figure~\ref{fig:process}. Hence, gaining a deeper understanding of the generation of adversarial examples with high transferability is essential to enhance model robustness.

Among the existing attack methods, the white-box attack strategy guarantees excellent attack performance by directly obtaining information about the target model but struggles to achieve high transfer attack performance, especially for models with defense mechanisms. To tackle this challenge, a series of methods have also been proposed to improve transferability for more practical black-box attacks. These methods can be mainly divided into three groups: gradient optimization~\citep{Long_Tao_LI_Lei_Zhang_2024, dong2017discovering,DBLP:journals/corr/abs-1908-06281,DBLP:conf/cvpr/Wang021}, data augmentation~\citep{xie2019improving,dong2019evading,wang2021admix}, and model augmentation~\citep{liu2016delving}. Furthermore, in most scenarios, integrating multiple attack methods can lead to improved attack performance.

Considering the aforementioned three attack perspectives, gradient optimization methods tend to be the most commonly used strategy. From the standpoint of optimization, the introduction of momentum~\citep{POLYAK19641} facilitates the accumulation of the attack gradient in the previous direction, achieving higher attack consistency to move the attack direction away from local optima. Most existing optimization methods, such as MI-FGSM~\citep{dong2017discovering}, NI-FGSM~\citep{DBLP:journals/corr/abs-1908-06281}, and VT~\citep{DBLP:conf/cvpr/Wang021}, are built upon the momentum. Therefore, we concentrate our analysis on momentum, phenomena of which remain prevalent in most circumstances. Although momentum can improve the effectiveness of attacks, we discover the \textbf{gradient elimination} phenomenon still exists during the forward attack process, i.e. the momentum fails to fully converge during the first few iterations leading to inaccurate attack directions, which may inhibit the transferability. Simultaneously, traditional iterative attacks (e.g. MI-FGSM, NI-FGSM) limit the optimization of generated adversarial examples within a very limited data distribution by using a small step size, increasing the likelihood of the attacks converging to local optima. Besides, a direct scaling of the step size may cause the attack easily falling into overfitting, as shown in Figure~\ref{fig:scale}.

To tackle the aforementioned challenges, our study commences by validating the association between gradient consistency and attack efficacy and scrutinizes the phenomenon of gradient elimination. We subsequently propose global momentum initialization to suppress gradient elimination and locate better global optimum over a larger data distribution during pre-convergence. Concretely, we undertake gradient pre-convergence in advance of the attack process to expedite momentum convergence, thus suppressing gradient elimination. Besides, we employ global search in the pre-convergence phase to help the initial momentum converge in a more effective direction. The generated adversarial examples under different methods are shown in Figure~\ref{fig:example}. We visualize the attack process in Figure~\ref{attack} for our method as well as the conventional method. The experimental outcomes demonstrate that our method yields a noteworthy improvement in the attack success rate, even against advanced defense mechanisms, exhibiting a 6.4\% increase in success rate over the current state-of-the-art method. Eventually, our approach attains an average attack success rate of 95.4\%. In summary, the main contributions are as follows:

$\bigcdot$ We verify the relationship between gradient consistency and attack performance from both theoretical and experimental perspectives. Based on the proposed gradient consistency, we determine the gradient elimination problem in the current attack methods.

$\bigcdot$ To suppress gradient elimination, we use pre-converg-ence as well as global search to improve gradient consistency. Further, we propose Global Momentum Initialization (GI), a novel and efficient attack algorithm, to help the attack find the global optimum.

$\bigcdot$ Our method can be seamlessly combined with any existing gradient-based attack method. Empirical experiments show that our method can significantly outperform existing state-of-the-art methods across diverse attack settings. Moreover, in addition to the image attack, our method be generalized to video attacks which have higher dimension.

The rest of the paper is organized as follows: In Section~2, we first provide an overview of gradient optimization and input transformation attack strategies, followed by a brief review of the main defense mechanisms. Section 3 delves into the notion of gradient consistency, based on which we further elucidate the importance of global momentum initialization (a priori momentum knowledge) in improving the efficiency of attacks. We also introduce our new strategy and provide the corresponding algorithmic framework. In Section 4, we present experimental results of black-box attacks in different settings. The experimental results of the attacks are presented for single model, multi-model, advanced defense methods, cross-structure model, and video model. To further analyse our approach, we conduct an ablation study and analyse the experimental results accordingly. Finally, in Section 5, we summarise our study and discuss the impact.

\begin{figure*}[h]
\begin{center}
	\subfloat[]{\label{fig:scale}\includegraphics[width = 0.4\textwidth]{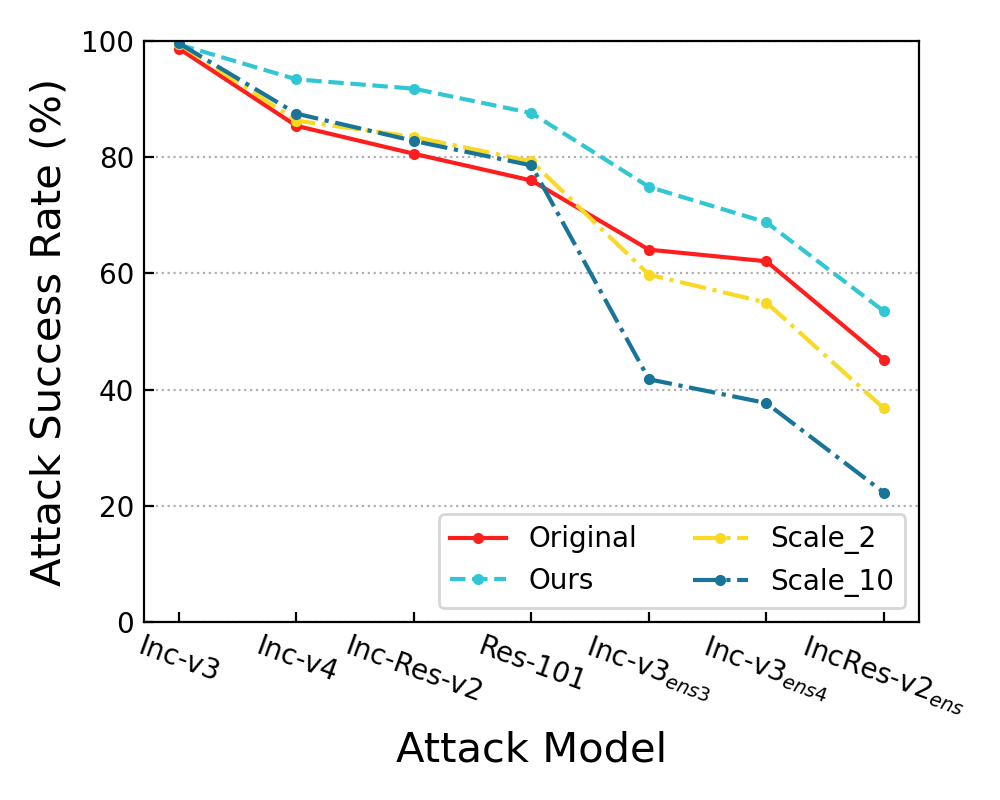}}
	  \hspace{-1mm}
	\subfloat[]{\label{fig:example}\includegraphics[width = 0.4\textwidth]{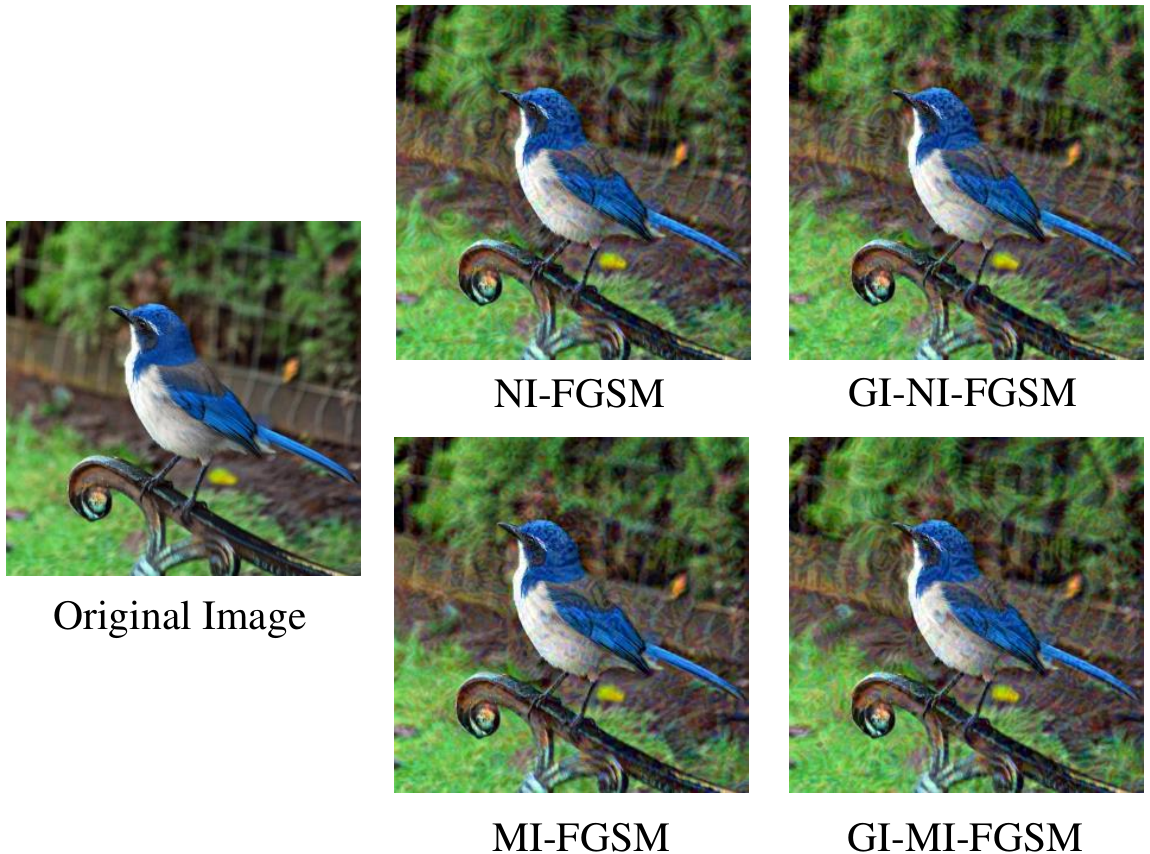}}
\end{center}
\caption{(a) shows the effect of the transfer attack at different step sizes. scale\_2 and scale\_10 show the results of the attack when the step size is enlarged by two times and ten times, respectively. (b) shows the adversarial examples under different attack methods. All the adversarial examples are generated with Inc-v3.}
\end{figure*}

\begin{figure}[h]
\begin{center}
\includegraphics[width=8.3cm,height=5.6cm]{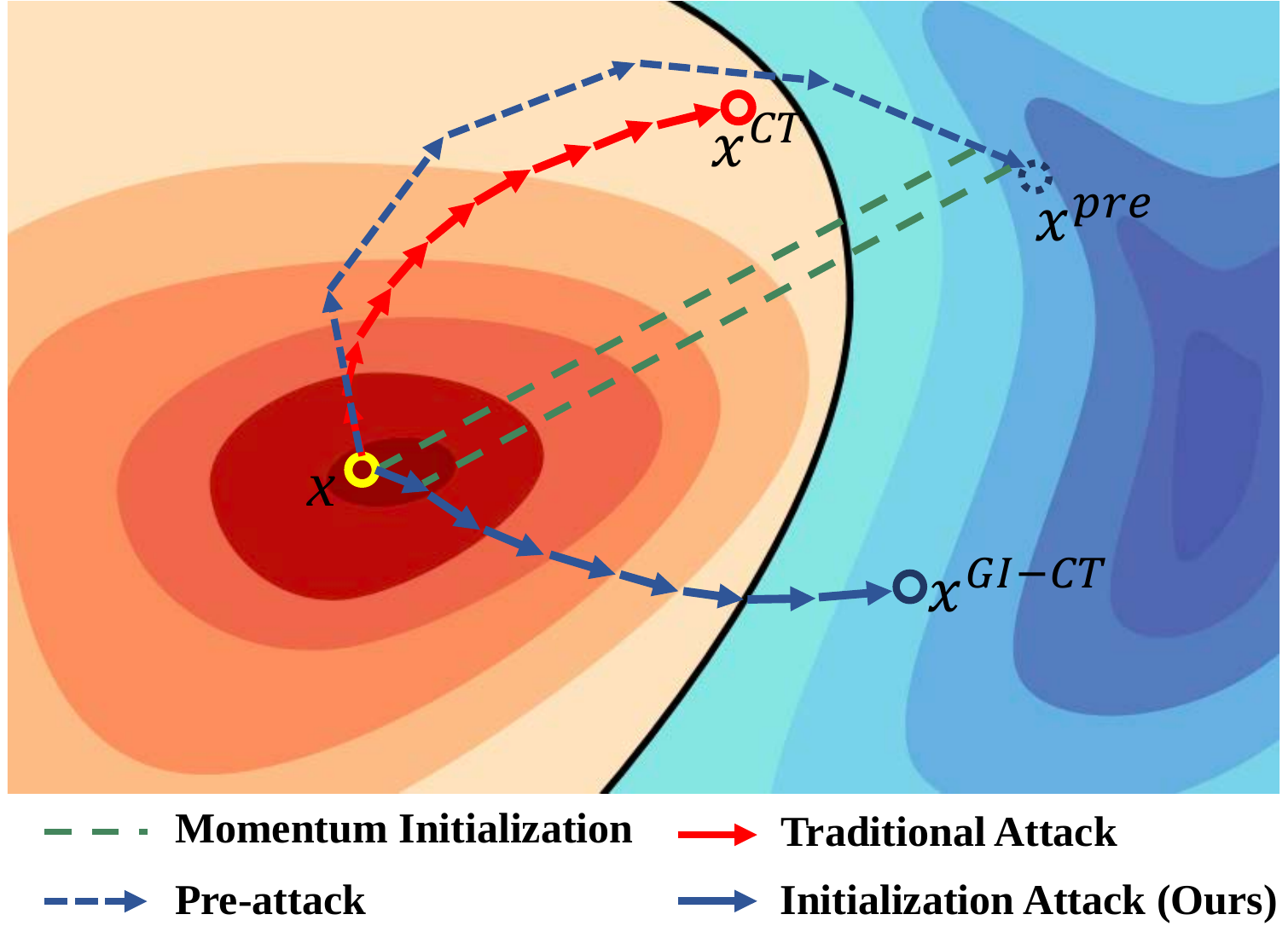}
\end{center}
\caption{Visualisation of the attack process. CT~\citep{DBLP:journals/corr/abs-1908-06281} represents the combination of three transformation methods: DIM~\citep{xie2019improving}, TIM~\citep{dong2019evading}, SIM~\citep{DBLP:journals/corr/abs-1908-06281}. $x$ represents the original data point and $x^{CT}$ represents the adversarial example obtained under original CT-FGSM. $x^{pre}$ and $x^{GI-CT}$ denote the overall process of the pre-attack as well as the formal attack.}
\label{attack}
\vspace{-14pt}
\end{figure}
\section{Related Work}

In this section, we introduce some mainstream attack algorithms from the perspectives of gradient optimization and input transformation. We also provide a brief overview of related advanced defense methods.

\subsection{Gradient Optimization Attacks}
\indent\textbf{Fast Gradient Sign Method (FGSM)}~\citep{fgsm}. FGSM generates an adversarial example for only one step with the aim of maximizing the loss function:
\begin{equation*}
    x^{adv} = x^{clean} + \epsilon \cdot {\rm sign}(\nabla_x J(x^{clean}, y)),
\end{equation*}
where sign(·) represents the sign function.

\textbf{Iterative Fast Gradient Sign Method (I-FGSM)}~\citep{kurakin2018adversarial}. In contrast to FGSM, I-FGSM divides one iteration into multiple small steps:
\begin{equation*}
        x^{adv}_{t+1} = x^{adv}_t + \alpha \cdot  {\rm sign}(\nabla_x J(x^{adv}_t, y)),
\end{equation*}
where $\alpha$ = $\epsilon / T$ is the step size of each attack and $T$ is the number of attack steps.

\textbf{Momentum Iterative Fast Gradient Sign Method (MI-FGSM)}~\citep{dong2017discovering}. Compared with white-box attacks, MI-FGSM integrates the momentum knowledge into attack to help the attack direction jump out of the local optimum:
\begin{equation}
\label{mi}
\begin{split}
            g_{t+1} = \mu \cdot  g_t + \frac{\nabla_x J(x^{adv}_t, y)}{||\nabla_x J(x^{adv}_t, y)||_1},
    \\x^{adv}_{t+1} = x^{adv}_t + \alpha \cdot {\rm sign}(g_{t+1}),
\end{split}
\end{equation}
where $\mu$ is the decay factor and $g_0$ = 0.

\textbf{Nesterov Iterative Fast Gradient Sign Method (NI-FGSM)}~\citep{DBLP:journals/corr/abs-1908-06281}. NI-FGSM uses Nesterov’s accelerated gradient~\citep{nesterov1983method} $x_t^{adv}+\alpha \cdot \mu \cdot g_t$ to replace $x_t^{adv}$ in Eq.~(\ref{mi}):
\begin{equation}
\label{ni}
\begin{split}
            g_{t+1} = \mu \cdot  g_t + \frac{\nabla_x J(x_t^{adv}+\alpha \cdot \mu \cdot g_t, y)}{||\nabla_x J(x_t^{adv}+\alpha \cdot \mu \cdot g_t, y)||_1},
    \\x^{adv}_{t+1} = x^{adv}_t + \alpha \cdot {\rm sign}(g_{t+1}),
\end{split}
\end{equation}
This look-ahead process allows attack to move further away from the local optimum, resulting in better attack performance.

\textbf{Varience Tuning Method (VT)}~\citep{DBLP:conf/cvpr/Wang021}. VT performs gradient sampling in the domain of data points in each iteration, which allows for the adjustment of the gradient direction with variance knowledge, thereby forming a more accurate and effective attack direction. 

Prior optimization methods have primarily focused on identifying a more optimal attack direction, while we present the first theoretical and experimental analysis from the perspective of gradient consistency and identify the gradient elimination issue. To solve the existing issue, we propose global momentum initialization to find a more effective attack direction. We hope that our proposed method can make up for the deficiencies of most of the existing gradient-based attack methods and provide a new optimization perspective for transfer attacks.

\subsection{Input Transformations Attacks}
\noindent 

\begin{figure*}[t]
\begin{center}
	\subfloat[]{\label{fig:sub1}\includegraphics[width = 0.31\textwidth]{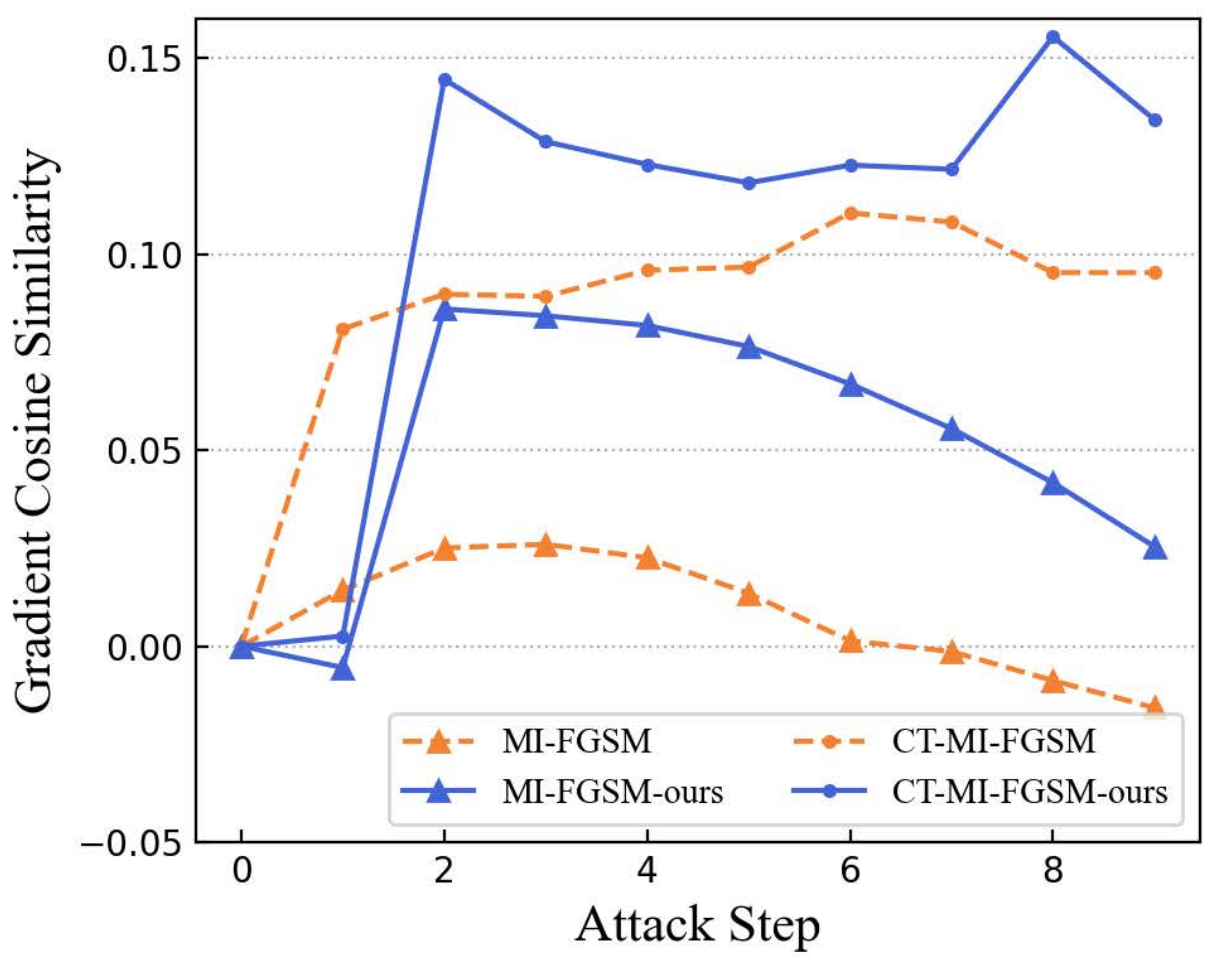}}
	\hfill
	\subfloat[]{\label{fig:sub2}\includegraphics[width = 0.305\textwidth]{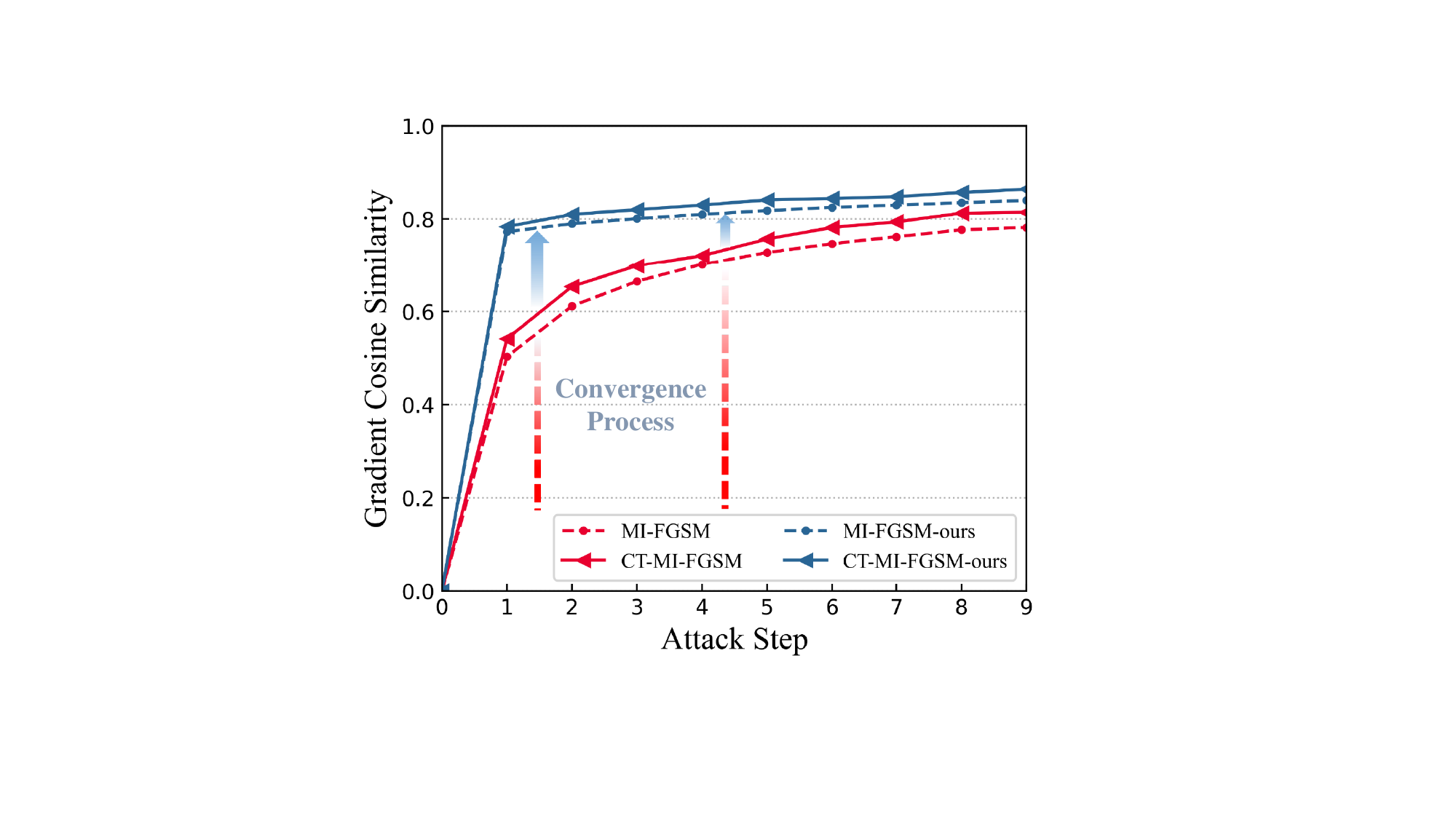}}
	\hfill
	\subfloat[]{\label{fig:sub3}\includegraphics[width = 0.32\textwidth]{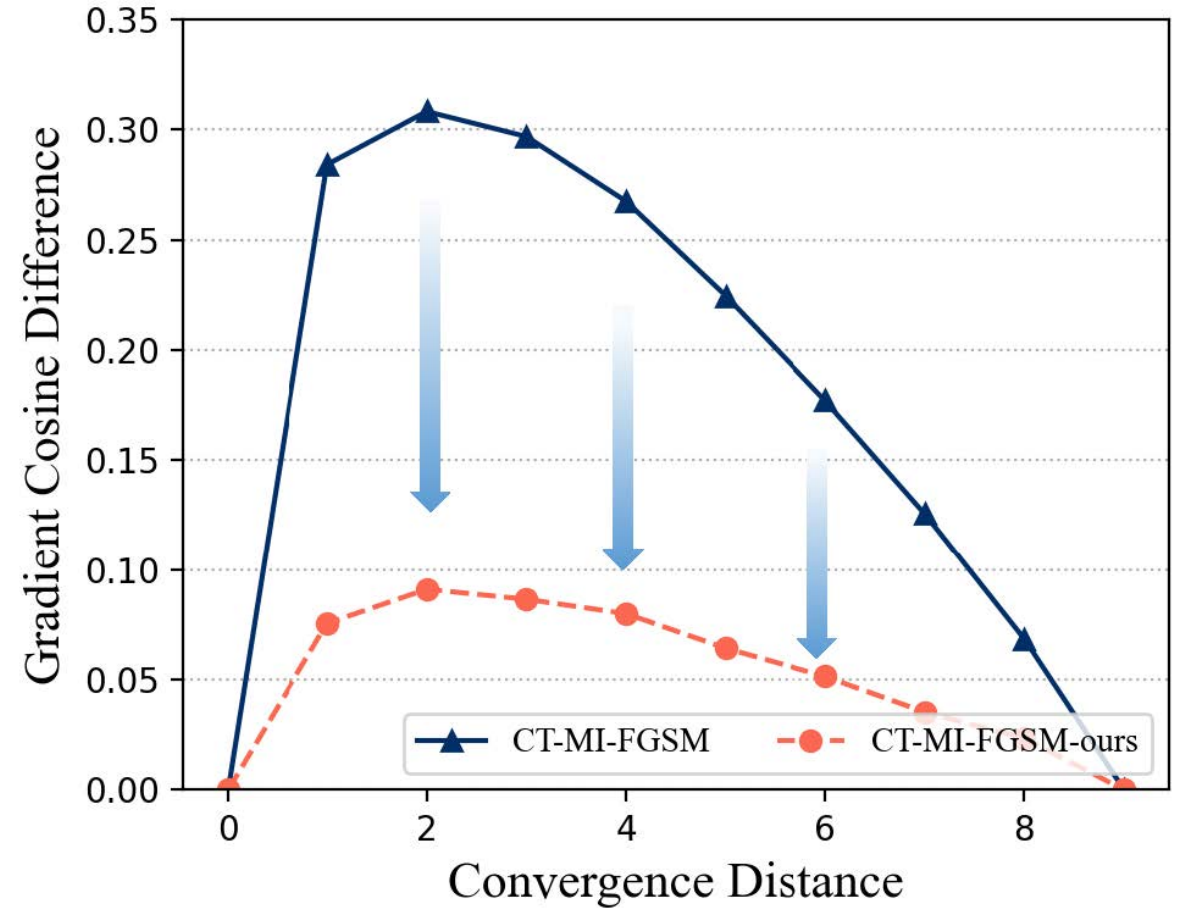}} 
\end{center}
\vspace{-14pt}
\caption{Analysis of gradient consistency (cosine similarity) between iterative attacks. (a) represents the gradient consistency before momentum accumulation; (b) represents the gradient consistency between iterations after momentum accumulation; (c) shows the gradient similarity analysis between the $1^{st}$, $10^{th}$ iteration and other iterations. For instance, when the convergence distance is 2, the gradient consistency between the $1^{st}$ and $3^{rd}$ rounds is nearly 0.3 smaller than that between the $10^{th}$ and $8^{th}$ rounds. However, with our approach, the difference is less than 0.1. The reduced difference suggests that the degree of consistency between the gradients in the $1^{st}$ round and those in the other rounds is very similar to the consistency between the gradients in the $10^{th}$ round and those in the other rounds. This implies a more adequate convergence.}
\label{grad_cos}
\vspace{-14pt}
\end{figure*}

\textbf{Diverse Input Method (DIM)}~\citep{xie2019improving}. DIM performs data augmentation using random scaling and padding operations with a certain probability to alleviate overfitting in adversarial attacks. This strategy provides valuable insights for enhancing transferability.

\textbf{Translation-Invariant Method (TIM)}~\citep{dong2019evading}. TIM applies convolution transformation of the Gaussian kernel $W$ to the gradient knowledge, resulting in a substantial enhancement of the attack performance, especially against adversarial-trained models.
\begin{equation*}
    x^{adv}_{t+1} = x^{adv}_t + \alpha \cdot  {\rm sign}(W \cdot \nabla_x J(D(x^{adv}_t), y)).
\end{equation*}

\textbf{Scale-Invariant Method (SIM)}~\citep{DBLP:journals/corr/abs-1908-06281}. SIM exploits the scaling invariance of the image and fuses the gradient knowledge of scaled images to improve the transferability. The new optimization objective after scaling is:
\begin{equation*}
    \mathop{\mathrm{argmax}}\limits_{x_{adv}}
    \frac{1}{n} \sum_{i=0}^n J(S_i(x^{adv}), y),
\end{equation*}
where scaling factor $S_i(x) = \frac{x}{2^i}$ and $n$ is the scale number.

~\citep{DBLP:journals/corr/abs-1908-06281} propose Composite Transformation (CT) obtained by combining DIM, TIM, SIM can be one of the most strongest input transformation methods. Our proposed approach, global momentum initialization, can be easily integrated with all input transformation methods.

\subsection{Advanced Adversarial Defense}
Many defense methods have been proposed in order to suppress the aggressiveness of adversarial examples against neural networks. The mainstream of adversarial defenses is divided into adversarial training as well as adversarial example preprocessing. Adversarial training methods~\citep{fgsm, DBLP:conf/iclr/KurakinGB17,DBLP:conf/iclr/MadryMSTV18} enhance the training data by adversarial examples, thereby refining the model decision boundaries to improve adversarial robustness. The adversarial example preprocessing method focuses on purifying the perturbations of adversarial examples as much as possible by some data enhancement means~\citep{guo2017countering,  DBLP:conf/icml/CohenRK19}, e.g. compression or smoothing.

\section{Methodology}
Given the original image $x^{clean}$ and the corresponding label $y$, the adversary has to find the perturbation $\delta$ to generate adversarial examples $x^{adv}$ to fool the model. To ensure the imperceptibility of the attack, we guarantee  $||x^{clean} - x^{adv}||_\infty \leq \epsilon$. Assume that $f(x^{adv})$  represents the output of the model with input $x$ and the attack is to maximize the loss $J(f(x^{adv}),y)$ (e.g. the cross-entropy loss or margin loss~\citep{carlini2017towards}), so the attack goal can be described as:
\begin{equation*}
    \mathop{\mathrm{argmax}}\limits_{x_{adv}}(J(f(x^{adv}),y)), \quad  s.t. ||x^{clean} - x^{adv}||_\infty \leq \epsilon.
\end{equation*}
\subsection{Gradient Elimination}
Among the existing gradient optimization methods, MI-FGSM is considered an important baseline. This method improves attack performance by introducing momentum to help accumulate gradients, resulting in more consistent attack directions. Despite the benefits of momentum in enabling the attack to escape from a local optimum, a pertinent question arises as to \textbf{whether early gradient attacks may have suboptimal attack performance due to a lack of momentum accumulation, given that momentum accumulation is a time-series process}. Since most optimization-based methods build upon the momentum of MI-FGSM, our analysis revolves around MI-FGSM, and can be extended to other optimization methods such as NI-FGSM. To delve deeper into the aforementioned issue, we leverage \textbf{gradient consistency}, i.e., the degree of gradient similarity during the attack, to investigate the correlation between attack gradient and attack optimization objective. We first develop a corollary of the numerical and theoretical relationship among different gradients.

Assuming that the gradient knowledge obtained in the first round of attack is $g_1$, then the direction of the first round of attack is $\mathrm{sign}(g_1)$, and the resulting difference of loss is:
\begin{equation}
\label{eq2}
    \Delta J_1=J(f(x+\alpha \cdot {\rm sign}(g_1)),y)-J(f(x),y).
\end{equation}
Since the step size $\alpha$ is very small with respect to $x$, we can treat $\alpha \cdot {\rm sign}(g_1)$ as a minimal value, i.e. $\alpha \cdot {\rm sign}(g_1) \rightarrow 0$. From the Taylor expansion we can know:
\begin{equation}
\label{eq3}
    f(x+\alpha \cdot{\rm sign}(g_1))=f(x)+\alpha \cdot {\rm sign}(g_1) \cdot \frac{\partial f}{\partial x}+O(\alpha^2).
\end{equation}
For simplicity, we denote $\alpha \cdot {\rm sign}(g_1) \cdot \frac{\partial f}{\partial x}+O(\alpha^2)$ as $P(x)$.
By substituting Eq.~(\ref{eq3}) into Eq.~(\ref{eq2}) and utilizing the Taylor expansion again, we can obtain:
\begin{equation}
\label{eq4}
    \begin{split}
        J(f(x),y)&=J(f(x+\alpha \cdot {\rm sign}(g_1))-P(x),y)\\
        &=J(f(x+\alpha \cdot {\rm sign}(g_1)),y)\\
        &-P(x) \cdot \frac{\partial J}{\partial f}+O(P^2(x)). 
    \end{split}
\end{equation}
Combining Eq.~(\ref{eq3}) and Eq.~(\ref{eq4}), since $(g_1 \neq 0)$, we can get:
\vspace{-2.4mm}
\begin{small}
\begin{equation}
\label{eq5}
    \begin{split}
    \Delta J_1
    =P(x) \cdot \frac{\partial J}{\partial f}=(\alpha \cdot {\rm sign}(g_1) \cdot \frac{\partial f}{\partial x}+O(\alpha^2)) \cdot \cfrac{\partial J}{\partial f}\\
    =\alpha \cdot {\rm sign}(g_1) \cdot g_1\approx  {\rm sign}(g_1) \cdot {\rm sign}(g_1) > 0 \quad(g_1 \neq 0).
    \end{split}
\end{equation}
% \vspace{-1.5mm}
\end{small}

Eq.~(\ref{eq5}) shows that when the first round gradient is attached to the input $x$, the loss change $\Delta J_1 >$  0, consistent with the fact that the loss increases during the attack process. Similarly, $\Delta J_n$ denotes that if the gradient of the $n_{th}$ iteration is applied to input $x$, the difference of loss changes to $\Delta J_n \approx  {\rm sign}(g_1) \cdot {\rm sign}(g_n)$. In this context, the input $x$ can be an adversarial example derived from any iteration, so we can test the consistency between gradients of any two rounds to predict the change in loss. In conjunction with the optimization objective, a more effective attack is one that gains greater losses with other rounds. Therefore, to improve attack performance, we should ensure relatively higher gradient consistency.

Based on the above theoretical analysis, we further explore the consistency of gradients under momentum from an experimental perspective. As shown in Figure~\ref{fig:sub1}, the cosine similarity among gradients prior to momentum convergence is nearly zero, meaning that there is a large randomness in attack directions. We argue that the low similarity and high randomness during the attack are the key factors in the overfitting of I-FGSM~\citep{kurakin2018adversarial}. On the other hand, Figure~\ref{fig:sub2} illustrates the gradient consistency of different rounds during the attack after the momentum accumulation, and it can be observed that after a few rounds of forward gradient convergence, the gradient consistency rapidly improves and reaches stability. This indicates that the momentum indeed helps the attack gradients reach more common directions, thus improving the attack performance. Simultaneously, when combined with existing data enhancement method CT~\citep{DBLP:journals/corr/abs-1908-06281}, the attack can explore better attack directions with higher attack consistency and attack performance.

However, there still exists inherent flaws in the current methodology. On the one hand, the time-series nature of the momentum convergence may lead to different accuracy of the information carried by the momentum at different time series. On the other hand, the accumulation of momentum within a very small data distribution is not conducive to finding a better global momentum and is more susceptible to overfitting. As Figure~\ref{fig:sub2} depicts, although momentum can accelerate the gradient to reach consistency, there remains a convergence process, which to a certain extent inhibits the attack performance improvement. We define this phenomenon where the gradient does not converge enough resulting in a less consistent attack direction as \textbf{gradient elimination}. 

To fully explore this phenomenon, in addition to exploring
neighboring gradient relations, we further investigate gradient consistency relations where there exists the same convergence distance. As shown in Figure~\ref{fig:sub3}, with the same convergence distance guaranteed, the forward unconverged gradient (e.g. the $1^{st}$ round) is significantly less consistent with the other rounds compared to the backward converged gradient (e.g. the $10^{th}$ round). This also suggests that the validity of the gradient knowledge is different for different rounds and that the validity is affected by momentum convergence, again confirming the phenomenon of gradient elimination.}

\begin{algorithm}[t]
\caption{Framework of GI-MI-FGSM}
\label{algo}
\textbf{Input: }A classifier $f$ with fixed parameters $\theta$ and loss function $J$, number of iterations $T$, maximum perturbation $\epsilon$, input images $x$, Pre-convergence Iterations $P$; Global Search Factor $S$ 

\textbf{Output: }An adversarial example $x^{adv}$;\\ %输出\begin{algorithm}[H]
\vspace{-5pt}
\begin{algorithmic}
\STATE $g_0 = 0$; $x_0^{adv} = x$; $\alpha = \frac{\epsilon}{T}$\;
\FOR{$t = 0$ to $P-1$}
    \STATE $g_{t+1} = \mu \cdot g_t + \frac{\nabla_x J(x^{adv}_t, y)}{||\nabla_x J(x^{adv}_t, y)||_1}$
    \STATE $x^{adv}_{t+1} = \text{Clip}(x^{adv}_t + S \cdot \alpha \cdot \text{sign}(g_{t+1}))$\;
\ENDFOR
\STATE Set global momentum initialization $g_0 = g_P$\;
\FOR{$t = 0$ to $T-1$}
    \STATE $g_{t+1} = \mu \cdot g_t + \frac{\nabla_x J(x^{adv}_t, y)}{||\nabla_x J(x^{adv}_t, y)||_1}$
    \STATE $x^{adv}_{t+1} = \text{Clip}(x^{adv}_t + \alpha \cdot \text{sign}(g_{t+1}))$\;
\ENDFOR
\STATE $x^{adv} = x^{adv}_T$\;
\RETURN $x^{adv}$
    
\end{algorithmic}
\end{algorithm}

\subsection{Global Momentum Initialization}
To address the challenge of gradient elimination, we introduce a pre-convergence attack strategy aimed at facilitating a more rapid convergence, thereby enhancing the effectiveness of the attack. Furthermore, we incorporate a global search mechanism during the pre-convergence phase to overcome the limitations associated with conventional iterative methods that generate gradients within a confined data distribution.

Specifically, our strategy involves pre-converging the momentum prior to the formal iterative attack, without exerting any substantive impact on the image through this process. We initiate  $P$ rounds of momentum exploration, using the obtained momentum knowledge as the initialised momentum. By means of this pre-convergence, we are able to substantially suppress the gradient elimination phenomenon. As shown in Figure~\ref{fig:sub2} and~\ref{fig:sub3}, the consistency of the forward gradient after pre-convergence is substantially improved, and the difference in gradient consistency between the forward unconverged and backward converged gradients and other rounds of gradients is significantly reduced.

Additionally, traditional iterative attacks use a small step size, causing the generated images after each round to be in a similar data distribution. This restricts the attack from exploring decision boundaries over a wider data range, making it more prone to getting stuck in local optimal solutions. ~\citep{gao2020patch} propose to scale up the step size, but this approach will have a direct impact on the images and easily cause overfitting of the attack. To address this issue, we add a \textbf{global search factor} to the pre-convergence process, i.e., we amplify the exploration step size during pre-convergence to help form a more global direction of initialised momentum. Namely, we perform gradient pre-convergence by:
\begin{gather}
g_{t+1} = \mu \cdot  g_t + \frac{\nabla_x J(x^{adv}_t, y)}{||\nabla_x J(x^{adv}_t, y)||_1},\nonumber \\
x^{adv}_{t+1} = {\rm Clip} (x^{adv}_t + S \cdot  \alpha 
\cdot  {\rm sign}(g_{t+1})),\nonumber
\end{gather}
where $S$ represents our global search factor and Clip($\cdot$) operation ensures the $l_\infty$ constraint. The procedure of the attack based on global momentum initialization can be summarised as Algorithm~\ref{algo}.

To further illustrate the efficacy of global momentum initialization in mitigating gradient elimination, we visualize the attack process in Figure~\ref{attack} for different methods. Since the direction of the attack gets progressively closer to the decision boundary during the optimization, traditional iterative methods may easily suffer from gradient elimination in the forward unconverged attack, i.e., they can only get closer to the decision boundary by a minor distance. For this reason, with the exploration of pre-attack and global momentum initialization, the attack strategy can search for a better attack direction in advance and ensure the consistency of the attack process to successfully cross the decision boundary.

\section{Experiment}
\subsection{Experiment Setup}

\textbf{Dataset.} 
Followed by NIPS'17 Competition~\citep{kurakin2018adversarial}, we randomly select 1000 images from the ILSVRC 2012 validation set~\citep{ilsvrc},  correctly classified and belonging to different categories as our original inputs.

\noindent\textbf{Models.}
We consider four normally trained models, including Inception-v3 (Inc-v3)~\citep{inceptionv2}, Inception-v4 (Inc-v4)~\citep{szegedy2017inception}, Inception-Resnet-v2 (IncRes-v2) and Resnet-101 (Res-101)~\citep{DBLP:conf/eccv/HeZRS16} as well as three adversarially trained models, namely ens3-adv-Inception-v3 (Inc-v3$_{ens3}$), ens4-adv-Inception-v3 (Inc-v3$_{ens4}$) and ens-adv-Inception-ResNet-v2 (IncRes-v2$_{ens}$)~\citep{tramer2017ensemble}. Instead of merely using Inception and ResNet as victim models, we also choose a series of recently popular models such as ViT~\citep{vit}, Swin-b~\citep{swint}, Mixer-mlp~\citep{mlp}, DeiT~\citep{deit}, and adversarially trained defense models as victim models. In addition, we adopt nine advanced defense methods to test the attack performance, e.g.  HGD~\citep{DBLP:conf/cvpr/LiaoLDPH018}, R\&P~\citep{xie2017mitigating}, NIPS-r3\footnote{\url{https://github.com/anlthms/nips-2017/tree/master/mmd}}, JPEG~\citep{guo2017countering}, FD~\citep{liu2019feature}, ComDefend~\citep{DBLP:conf/cvpr/JiaWCF19}, NRP~\citep{DBLP:conf/cvpr/NaseerKHKP20}, RS~\citep{DBLP:conf/icml/CohenRK19}, Bit-Red~\citep{DBLP:conf/ndss/Xu0Q18}. 

\noindent\textbf{Baselines.}
We follow the standard experimental setting of ~\citep{DBLP:conf/cvpr/Wang021,xie2019improving,wang2021admix}. For optimization methods, we regard MI-FGSM~\citep{dong2017discovering}, NI-FGSM~\citep{DBLP:journals/corr/abs-1908-06281} and the VT~\citep{DBLP:conf/cvpr/Wang021} as our baselines. Simultaneously, for data augmentation methods, we consider DIM~\citep{xie2019improving}, TIM~\citep{dong2019evading}, SIM, CT~\citep{DBLP:journals/corr/abs-1908-06281}, i.e., the integrated version of the former three methods. When combining our method with optimization and input transformation methods, we denote final methods as GI-VM(N)I-CT-FGSM, GI-M(N)I-CT-FGSM and GI-M(N)I-FGSM, respectively. Moreover, we have additionally selected several recent state-of-the-art methods as baselines such as SIA~\citep{Wang_2023_ICCV}, Admix~\citep{wang2021admix}, EMI~\citep{DBLP:journals/corr/abs-2103-10609}, SSA~\citep{ssa}, PI-FGSM~\citep{gao2020patch}.

\noindent\textbf{Hyper-parameters.}
We follow the standard attack settings of ~\citep{DBLP:conf/cvpr/Wang021,xie2019improving,wang2021admix}, where we set the maximum perturbation $\epsilon$ to 16, the number of iteration rounds $T$ to 10, the iteration step size $\alpha$ to 1, and the decay factor $\mu$ to 1 for MI-FGSM and NI-FGSM. As for input transformation methods, we set the transformation probability to 0.5 for DIM. We adopt 7$\times$7 Gaussian kernel for TIM and the scale number for SIM equals 5. For our method, we set pre-convergence iterations $P$ to 5 and global search factor $S$ to 10.

\begin{table*}[t]
\renewcommand{\arraystretch}{0.9} 
\begin{center}
\scalebox{0.73}{
\begin{tabular}{@{}c|c|cccccccc@{}}
\toprule
Model
                           & Attack    & Inc-v3          & Inc-v4          & IncRes-v2      & Res-101        & Inc-v3$_{ens3}$    & Inc-v3$_{ens4}$    & IncRes-v2$_{ens}$  & Avg.       \\ \midrule
\multirow{4}{*}{Inc-v3}    & MI-FGSM    & \textbf{100.0*} & 44.4            & 41.5           & 34.7           & \textbf{14.5} & 12.4          & 6.0           & 36.2          \\
                           & GI-MI-FGSM & \textbf{100.0*} & \textbf{54.1}   & \textbf{51.9}  & \textbf{43.8}  & 14.3          & \textbf{13.4} & \textbf{6.6}  & \textbf{40.6} \\ \cmidrule(l){2-10} 
                           & NI-FGSM    & \textbf{100.0*} & 52.1            & 49.9           & 42.6           & \textbf{13.7} & \textbf{13.9} & 6.0           & 39.7          \\
                           & GI-NI-FGSM & \textbf{100.0*} & \textbf{58.7}   & \textbf{56.0}  & \textbf{47.5}  & 13.4          & 12.0          & \textbf{6.8}  & \textbf{42.1} \\ \midrule
\multirow{4}{*}{Inc-v4}    & MI-FGSM    & 56.3            & 99.7*           & 46.7           & 41.3           & \textbf{16.4} & 14.8          & 7.6           & 40.4          \\
                           & GI-MI-FGSM & \textbf{68.6}   & \textbf{100.0*} & \textbf{57.0}  & \textbf{51.0}  & 16.1          & \textbf{15.8} & \textbf{7.7}  & \textbf{45.2} \\ \cmidrule(l){2-10} 
                           & NI-FGSM    & 63.3            & \textbf{100.0*} & 51.1           & 45.9           & 15.1          & 14.4          & 7.0           & 42.4          \\
                           & GI-NI-FGSM & \textbf{74.7}   & \textbf{100.0*} & \textbf{62.4}  & \textbf{53.3}  & \textbf{16.8} & \textbf{16.2} & \textbf{8.1}  & \textbf{47.4} \\ \midrule
\multirow{4}{*}{\makecell{IncRes\\-v2}} & MI-FGSM    & 58.7            & 50.9            & 98.1*          & 45.1           & 21.6          & 17.2          & 11.5          & 43.3          \\
                           & GI-MI-FGSM & \textbf{73.9}   & \textbf{64.3}   & \textbf{98.5*} & \textbf{57.2}  & \textbf{23.0} & \textbf{17.7} & \textbf{12.3} & \textbf{49.6} \\ \cmidrule(l){2-10} 
                           & NI-FGSM    & 61.6            & 54.5            & 99.1*          & 45.5           & 20.1          & 16.0          & 9.5           & 43.8          \\
                           & GI-NI-FGSM & \textbf{77.1}   & \textbf{68.3}   & \textbf{99.5*} & \textbf{58.8}  & \textbf{23.8} & \textbf{18.0} & \textbf{11.7} & \textbf{51.0} \\ \midrule
\multirow{4}{*}{\makecell{Res\\-101}}   & MI-FGSM    & 58.8            & 50.8            & 50.8           & 99.1*          & 24.3          & \textbf{22.0} & \textbf{12.9} & 45.5          \\
                           & GI-MI-FGSM & \textbf{71.1}   & \textbf{64.6}   & \textbf{64.2}  & \textbf{99.3*} & \textbf{26.2} & 21.2          & 12.4          & \textbf{51.3} \\ \cmidrule(l){2-10} 
                           & NI-FGSM    & 64.3            & 59.4            & 55.9           & 99.4*          & 24.1          & \textbf{21.2} & \textbf{12.3} & 48.1          \\
                           & GI-NI-FGSM & \textbf{74.7}   & \textbf{68.6}   & \textbf{65.9}  & \textbf{99.5*} & \textbf{26.9} & 21.1          & 12.0          & \textbf{52.7} \\ \bottomrule
\end{tabular}}
\end{center}
\vspace{-8pt}
\caption{Attack success rates (\%) of seven models using optimization method merely. The adversarial examples are crafted by Inc-v3, Inc-v4, IncRes-v2, Res-101 respectively. $*$ indicates the white box attack setting.}
\label{tab1}
\vspace{-8pt}
\end{table*}

\subsection{Attack with Optimization Methods}
Initially, we evaluate the performance of our method with the optimization method only, i.e., we combine our method to NI-FGSM and MI-FGSM respectively without considering the input transformation methods. Here we use the attack success rate, the proportion of images misclassified by the attack model as our metric. During the attack, we use four normally trained models as surrogate models to generate adversarial examples iteratively and then transfer them to all models, the results of which are shown in Table~\ref{tab1}.

\begin{table*}[]
\centering
\scalebox{0.7}{
\begin{tabular}{@{}c|c|cccccccc@{}}
\toprule
    Model                       & Attack        & Inc-v3         & Inc-v4         & IncRes-v2      & Res-101        & Inc-v3$_{ens3}$    & Inc-v3$_{ens4}$    & IncRes-v2$_{ens}$  & Avg.       \\ \midrule
\multirow{8}{*}{Inc-v3}    & NI-CT-FGSM     & 99.5*          & 84.5           & 79.0           & 72.9           & 57.2          & 54.8          & 40.3          & 69.7          \\
                           & GI-NI-CT-FGSM  & \textbf{99.8*} & \textbf{93.7}  & \textbf{91.0}  & \textbf{87.1}  & \textbf{70.1} & \textbf{65.9} & \textbf{50.0} & \textbf{79.7} \\ \cmidrule(l){2-10} 
                           & VNI-CT-FGSM    & 99.0*          & 90.1           & 87.6           & 84.6           & 78.6          & 76.5          & 66.0          & 83.2          \\
                           & GI-VNI-CT-FGSM & \textbf{99.3*} & \textbf{94.5}  & \textbf{92.7}  & \textbf{89.7}  & \textbf{85.1} & \textbf{83.3} & \textbf{72.9} & \textbf{88.2} \\ \cmidrule(l){2-10} 
                           & MI-CT-FGSM     & 98.7*          & 85.4           & 80.6           & 76.0           & 64.1          & 62.1          & 45.2          & 73.2          \\
                           & GI-MI-CT-FGSM  & \textbf{99.4*} & \textbf{93.4}  & \textbf{91.8}  & \textbf{87.6}  & \textbf{74.9} & \textbf{68.8} & \textbf{53.5} & \textbf{81.3} \\ \cmidrule(l){2-10} 
                           & VMI-CT-FGSM    & 99.0*          & 88.5           & 85.7           & 82.4           & 77.3          & 75.9          & 62.8          & 81.7          \\
                           & GI-VMI-CT-FGSM & \textbf{99.5*} & \textbf{97.1}  & \textbf{95.9}  & \textbf{92.8}  & \textbf{88.0} & \textbf{86.5} & \textbf{74.9} & \textbf{90.7} \\ \midrule
\multirow{8}{*}{Inc-v4}    & NI-CT-FGSM     & 87.8           & 99.4*          & 82.5           & 75.9           & 65.8          & 62.6          & 51.3          & 75.0          \\
                           & GI-NI-CT-FGSM  & \textbf{95.0}  & \textbf{99.5*} & \textbf{93.4}  & \textbf{87.3}  & \textbf{76.7} & \textbf{71.2} & \textbf{59.5} & \textbf{83.2} \\ \cmidrule(l){2-10} 
                           & VNI-CT-FGSM    & 92.3           & 99.7*          & 89.2           & 84.9           & 79.9          & 77.1          & 70.5          & 84.8          \\
                           & GI-VNI-CT-FGSM & \textbf{97.6}  & \textbf{99.7*} & \textbf{92.6}  & \textbf{90.7}  & \textbf{88.1} & \textbf{86.2} & \textbf{79.8} & \textbf{90.7} \\ \cmidrule(l){2-10} 
                           & MI-CT-FGSM     & 87.2           & 98.6*          & 83.3           & 78.3           & 72.2          & 67.2          & 57.3          & 77.7          \\
                           & GI-MI-CT-FGSM  & \textbf{94.5}  & \textbf{99.5*} & \textbf{93.2}  & \textbf{87.8}  & \textbf{77.3} & \textbf{73.8} & \textbf{61.2} & \textbf{83.9} \\ \cmidrule(l){2-10} 
                           & VMI-CT-FGSM    & 89.7           & 98.8*          & 86.3           & 82.0           & 78.1          & 76.2          & 67.5          & 82.7          \\
                           & GI-VMI-CT-FGSM & \textbf{97.0}  & \textbf{99.8*} & \textbf{95.5}  & \textbf{91.7}  & \textbf{88.3} & \textbf{86.8} & \textbf{79.4} & \textbf{91.2} \\ \midrule
\multirow{8}{*}{\makecell{IncRes\\-v2}} & NI-CT-FGSM     & 90.2           & 87.0           & 99.4*          & 83.2           & 75.0          & 68.9          & 65.1          & 81.3          \\
                           & GI-NI-CT-FGSM  & \textbf{96.0}  & \textbf{95.3}  & \textbf{99.4*} & \textbf{92.0}  & \textbf{84.3} & \textbf{80.7} & \textbf{75.7} & \textbf{89.1} \\ \cmidrule(l){2-10} 
                           & VNI-CT-FGSM    & 92.9           & 90.6           & 99.0*          & 88.2           & 85.2          & 82.5          & 81.8          & 88.6          \\
                           & GI-VNI-CT-FGSM & \textbf{96.3}  & \textbf{97.3}  & \textbf{99.3*} & \textbf{95.4}  & \textbf{94.4} & \textbf{91.6} & \textbf{91.0} & \textbf{95.0} \\ \cmidrule(l){2-10} 
                           & MI-CT-FGSM     & 88.0           & 85.5           & 97.5*          & 81.6           & 76.0          & 71.5          & 70.2          & 81.5          \\
                           & GI-MI-CT-FGSM  & \textbf{95.3}  & \textbf{94.7}  & \textbf{99.0*} & \textbf{91.4}  & \textbf{85.6} & \textbf{82.1} & \textbf{78.4} & \textbf{89.5} \\ \cmidrule(l){2-10} 
                           & VMI-CT-FGSM    & 88.9           & 87.0           & 97.0*          & 85.0           & 83.4          & 80.5          & 79.4          & 85.9          \\
                           & GI-VMI-CT-FGSM & \textbf{96.9}  & \textbf{96.3}  & \textbf{99.4*} & \textbf{94.4}  & \textbf{92.6} & \textbf{91.0} & \textbf{89.9} & \textbf{94.4} \\ \midrule
\multirow{8}{*}{\makecell{Res\\-101}}   & NI-CT-FGSM     & 86.1           & 82.2           & 83.3           & 98.5*          & 70.0          & 68.5          & 54.6          & 77.6          \\
                           & GI-NI-CT-FGSM  & \textbf{94.9}  & \textbf{92.7}  & \textbf{93.2}  & \textbf{99.3*} & \textbf{83.4} & \textbf{78.3} & \textbf{68.7} & \textbf{87.2} \\ \cmidrule(l){2-10} 
                           & VNI-CT-FGSM    & 90.7           & 85.5           & 87.2           & 99.1*          & 82.6          & 79.7          & 73.3          & 85.4          \\
                           & GI-VNI-CT-FGSM & \textbf{96.3}  & \textbf{95.6}  & \textbf{95.5}  & \textbf{99.4*} & \textbf{92.4} & \textbf{90.5} & \textbf{85.5} & \textbf{93.6} \\ \cmidrule(l){2-10} 
                           & MI-CT-FGSM     & 86.5           & 81.8           & 83.2           & 98.9*          & 77.0          & 72.3          & 61.9          & 80.2          \\
                           & GI-MI-CT-FGSM  & \textbf{94.0}  & \textbf{92.2}  & \textbf{91.8}  & \textbf{99.3*} & \textbf{83.9} & \textbf{80.4} & \textbf{70.0} & \textbf{87.4} \\ \cmidrule(l){2-10} 
                           & VMI-CT-FGSM    & 86.9           & 84.2           & 86.4           & 98.6*          & 81.0          & 78.6          & 71.6          & 83.9          \\
                           & GI-VMI-CT-FGSM & \textbf{95.8}  & \textbf{95.3}  & \textbf{94.9}  & \textbf{99.3*} & \textbf{92.3} & \textbf{89.8} & \textbf{85.2} & \textbf{93.2} \\ \bottomrule
\end{tabular}}
\caption{Attack success rates (\%) of seven black models using both optimization methods and transformation methods. The adversarial examples are crafted by Inc-v3, Inc-v4, IncRes-v2, Res-101 respectively. * indicates the white box attack setting.}
\label{tab2}
\vspace{-8pt}
\end{table*}

\begin{table*}[t]
\begin{center}
\scalebox{0.66}{
\begin{tabular}{@{}c|cccccccc@{}}
\toprule
Attack              & Inc-v3          & Inc-v4        & IncRes-v2     & Res-101       & Inc-v3ens3    & Inc-v3ens4    & IncRes-v2ens  & Average       \\ \midrule
PI-CT-MI-FGSM       & 99.4*           & 82.1          & 79.8          & 76.4          & 57.8          & 53.1          & 38.9          & 69.6          \\
GI-PI-CT-MI-FGSM    & \textbf{99.5*}  & \textbf{88.8} & \textbf{85.9} & \textbf{79.2} & \textbf{60.6} & \textbf{56.8} & \textbf{40.7} & \textbf{73.1} \\ \midrule
SSA-CT-MI-FGSM      & 99.8*           & 93.5          & 92.3          & 89.4          & 89.7          & 88.1          & 78.5          & 90.2          \\
GI-SSA-CT-MI-FGSM   & \textbf{100.0*} & \textbf{96.9} & \textbf{96.3} & \textbf{94.2} & \textbf{92.4} & \textbf{91.3} & \textbf{83.2} & \textbf{93.5} \\ \midrule
Admix-CT-MI-FGSM    & 99.8*           & 89.2          & 86.4          & 82.7          & 72.5          & 67.3          & 51.5          & 78.5          \\
GI-Admix-CT-MI-FGSM & \textbf{100.0*} & \textbf{96.8} & \textbf{94.5} & \textbf{89.0} & \textbf{79.8} & \textbf{74.6} & \textbf{57.0} & \textbf{84.5} \\ \midrule
EMI-TIDIMI-FGSM     & 99.6*           & 86.4          & 83.0          & 76.1          & 53.7          & 51.0          & 35.0          & 69.3          \\
GI-EMI-TIDIMI-FGSM  & \textbf{99.7*}  & \textbf{93.7} & \textbf{92.3} & \textbf{85.6} & \textbf{62.4} & \textbf{56.4} & \textbf{40.5} & \textbf{75.8} \\ \midrule
SIA-MI-FGSM         & 99.6            & 90.5          & 88.6          & 82.7          & 75.5          & 69.5          & 54.7          & 80.2          \\
GI-SIA-MI-FGSM      & \textbf{100.0*} & \textbf{96.4} & \textbf{94.9} & \textbf{90.9} & \textbf{85.2} & \textbf{80.1} & \textbf{65.6} & \textbf{87.6} \\ \bottomrule
\end{tabular}}
\end{center}
\vspace{-8pt}
\caption{Attack success rate (\%) of more state-of-the-art methods and the combination with our method. * represents the white-box attack setting. Adversarial examples are generated by Inc-v3.}
\label{table:ssa}
\end{table*}

According to the results, we can see that for the four normally trained models, our method improves the black-box attack transferability by a large margin while improving the white-box performance as well, e.g., for the adversarial examples generated by IncRes-v2 and transferred to Inc-v3, the attack success rate increases from 58.7\% and 61.6\% to 73.9\% and 77.1\%, respectively. However, at the same time, we also notice that the improvement for the adversarial-trained models is not as significant as that for the normally trained models. We attribute this to the inherent low transferability of the adversarially trained model, which results in the original attack direction not being able to explore a better global optimal direction even after better optimization. This phenomenon will be eliminated when combined with the input transformation methods. Details are available in Section~\ref{section:ct}.

\subsection{Attack with State-of-the-art}
\label{section:ct}
Further, we verify the performance of GI with both the optimization and input transformation methods. We follow the experimental setup CT in \citep{DBLP:journals/corr/abs-1908-06281} and then combine them with the gradient optimization methods: MI-FGSM, NI-FGSM and VT respectively to test the performance. It is crucial to note that the VT entails a non-negligible time and space cost concerning attack performance. To ensure a fair comparison, we test attack performance under both M(N)I-CT-FGSM and VM(N)I-CT-FGSM. The final results are shown in Table~\ref{tab2}. In the M(N)I-CT-FGSM condition, our proposed method demonstrates a notable enhancement in the attack success rate, achieving an average improvement of 8.1\%. Moreover, our approach achieves a comparable attack performance to the previous state-of-the-art method VM(N)I-CT-FGSM, all the while requiring less than \textbf{one-tenth of the time} to execute. In particular, the final method, GI-M(N)I-VT-FGSM, can achieve an average attack success rate of 88.2\% $\sim$ 95.0\%, which exceeds the previous state-of-the-art method VM(N)I-CT-FGSM by 7.6\% in average. 

In addition to the previously mentioned baselines, there still exits several popular state-of-the-art methods for enhancing transferability. ~\citep{Wang_2023_ICCV} devise the Structure Invariant Attack (SIA), employing various data augmentation techniques to enhance transferability. ~\citep{ssa} propose the use of Spectrum Simulation Attack (SSA) in the frequency domain space to improve attack performance. Admix~\citep{wang2021admix} obtains better gradients for the original image mixed with other images by data enhancement to improve the transferability. EMI~\citep{DBLP:journals/corr/abs-2103-10609} is looking ahead with the guidance of local gradient knowledge to get a better attack direction. PI-FGSM~\citep{gao2020patch}  considers projecting the clipped gradient knowledge to make full use of the gradient knowledge to improve attack performance. We also incorporate our method into these state-of-the-art methods to test the performance and the results are shown in Table~\ref{table:ssa}. For EMI, we utilize the official source code provided without the integration of CT. Indeed, in all instances, our approach exhibits a substantial improvement over the existing performance.

Aside from evaluating existing model structures, our experimentation also encompasses popular model frameworks like \textbf{ViT}, \textbf{DeiT}, \textbf{Mixer-mlp}, and \textbf{Swin-b}, along with several adversarial-trained models. For details of the experimental outcomes, please refer to Table~\ref{tab:vit}.

\begin{table}[t]
\begin{center}
\scalebox{0.6}{
\begin{tabular}{@{}c|ccccccc@{}}
\toprule
Attack         & ViT           & Swin-b        & Mixer-mlp     & DeiT          & Resnet-101$_{adv}$ & Swin-b$_{adv}$     & Avg           \\ \midrule
MI-CT-FGSM     & 72.0          & 83.4          & 82.9          & 81.8          & 25.7          & 20.6          & 61.1          \\
GI-MI-CT-FGSM  & \textbf{81.9} & \textbf{90.1} & \textbf{92.1} & \textbf{91.0} & \textbf{28.6} & \textbf{21.8} & \textbf{67.6} \\ \midrule
VMI-CT-FGSM    & 76.6          & 85.6          & 86.8          & 84.7          & 28.4          & 22.7          & 64.1          \\
GI-VMI-CT-FGSM & \textbf{87.2} & \textbf{92.3} & \textbf{94.0} & \textbf{94.2} & \textbf{30.9} & \textbf{25.2} & \textbf{70.6} \\ \bottomrule
\end{tabular}
}
\end{center}
\vspace{-8pt}
\caption{\footnotesize Attack success rates (\%) under different model architectures. The adversarial examples are crafted by the ensemble of Inc-v3, Inc-v4, IncRes-v2, Res-101 models.}
\label{tab:vit}
\vspace{-8pt}
\end{table}

\vspace{-6pt}
\subsection{Attack with Ensemble Models}
The preceding two subsections have effectively demonstrated the generality and high performance of our approach within the single-source white-box model setting. ~\citep{liu2016delving} have shown that the adversarial examples generated with the ensemble model are more transferable. We proceed to verify the performance of our approach under the ensemble attack scenario. The results of this evaluation are presented in Table~\ref{tab3}. The results indicate that our method yields a significant improvement in the effectiveness of the black-box attack while also enhancing the performance of the white-box attack to some extent under both CT and VT conditions. It is noteworthy that with the ensemble model method, our attack success rate reaches over 98.6\% on average, which means that we can basically achieve the performance of white-box attack. Furthermore, under the white-box attack setting, the attack success rate can reach 100\% in most circumstances.

\begin{table*}[t]
\begin{center}
\scalebox{0.7}{
\begin{tabular}{@{}c|cccccccc@{}}
\toprule
Attack      & Inc-v3          & Inc-v4          & Inc-Res-v2      & Res-101        & Inc-v3$_{ens3}$    & Inc-v3$_{ens4}$    & IncRes-v2$_{ens}$  & Avg.       \\ \midrule
MI-CT-FGSM  & 99.4*           & 99.0*           & 97.6*           & 99.7*           & 91.3          & 90.1          & 86.6          & 94.8          \\
GI-MI-CT-FGSM  & \textbf{100.0*} & \textbf{99.9*}  & \textbf{99.8*}  & \textbf{100.0*} & \textbf{98.0} & \textbf{97.2} & \textbf{95.0} & \textbf{98.6} \\ \midrule
NI-CT-FGSM  & \textbf{100.0*} & \textbf{100.0*} & 99.8*           & \textbf{100.0*} & 91.8          & 89.1          & 84.5          & 95.0          \\
GI-NI-CT-FGSM  & \textbf{100.0*} & 99.9*           & \textbf{100.0*} & \textbf{100.0*} & \textbf{98.2} & \textbf{97.4} & \textbf{95.4} & \textbf{98.7} \\ \midrule
VMI-CT-FGSM & 99.7*           & 99.1*           & 98.6*           & \textbf{100.0*} & 93.2          & 92.3          & 90.4          & 96.2          \\
GI-VMI-CT-FGSM & \textbf{100.0*} & \textbf{100.0*} & \textbf{100.0*} & \textbf{100.0*} & \textbf{98.9} & \textbf{98.6} & \textbf{97.4} & \textbf{99.3} \\ \midrule
VNI-CT-FGSM & 99.7*           & 99.6*           & 99.2*           & 99.9*           & 94.9          & 93.8          & 92.0          & 97.0          \\
GI-VNI-CT-FGSM & \textbf{100.0*} & \textbf{100.0*} & \textbf{100.0*} & \textbf{100.0*} & \textbf{98.9} & \textbf{98.5} & \textbf{97.6} & \textbf{99.3} \\ \bottomrule
\end{tabular}}
\end{center}
\vspace{-8pt}
\caption{Attack success rates (\%) of seven models using both optimization methods and input transformation methods. The adversarial examples are crafted by the ensemble of Inc-v3, Inc-v4, IncRes-v2, Res-101 models. * indicates the white box attack setting.}
\label{tab3}
\vspace{-8pt}
\end{table*}

\subsection{Evaluation on Advanced Defense Methods}
To further investigate the performance of our approach under advanced defense methods, we selected several official baselines from the NIPS 2017 competition~\citep{kurakin2018adversarial} for testing, such as HGD~\citep{DBLP:conf/cvpr/LiaoLDPH018}, R\&P~\citep{xie2017mitigating}, NIPS-r3, along with several classical methods for adversarial defense, such as JPEG~\citep{guo2017countering}, FD~\citep{liu2019feature}, ComDefend~\citep{DBLP:conf/cvpr/JiaWCF19}, NRP~\citep{DBLP:conf/cvpr/NaseerKHKP20}, RS~\citep{DBLP:conf/icml/CohenRK19}, Bit-Red~\citep{DBLP:conf/ndss/Xu0Q18}, to further test the performance of attacks under defense conditions. Here we test the attack performance with the integrated model and the single model (Inc-v3) respectively. The attack results for the integrated model are shown in Table~\ref{tab4}.

The results are impressive in that our attack method achieves an average attack success rate of 91.6\% $\sim$ 95.4\%, outperforming the previous VM(N)I-CT-FGSM method by 6.4\% $\sim$ 9.0\%, which also indicates the vulnerability and fragility of the existing defense methods.

\begin{table*}[t]
\centering
\scalebox{0.72}{
\begin{tabular}{@{}c|cccccccccc@{}}

\toprule
Attack      & HGD           & R\&P          & NIPS-r3       & JPEG          & FD            & ComDefend     & NRP           & RS            & Bit-Red       & Avg.       \\ \midrule
MI-CT-FGSM  & 90.8          & 88.1          & 87.6          & 92.1          & 88.5          & 90.4          & 76.4          & 69.3          & 76.4          & 84.4          \\
GI-MI-CT-FGSM  & \textbf{97.0} & \textbf{95.2} & \textbf{95.8} & \textbf{98.2} & \textbf{95.7} & \textbf{98.1} & \textbf{83.1} & \textbf{78.7} & \textbf{84.3} & \textbf{91.8} \\ \midrule
NI-CT-FGSM  & 90.7          & 86.6          & 86.5          & 92.8          & 89.7          & 91.3          & 69.8          & 64.2          & 72.2          & 82.6          \\
GI-NI-CT-FGSM  & \textbf{97.6} & \textbf{95.7} & \textbf{95.9} & \textbf{98.3} & \textbf{95.9} & \textbf{97.3} & \textbf{81.4} & \textbf{78.8} & \textbf{83.8} & \textbf{91.6} \\ \midrule
VMI-CT-FGSM & 92.8          & 90.4          & 89.9          & 93.4          & 91.2          & 92.2          & 83.3          & 76.9          & 80.4          & 87.8          \\
GI-VMI-CT-FGSM & \textbf{98.5} & \textbf{97.7} & \textbf{97.4} & \textbf{99.0} & \textbf{96.6} & \textbf{98.3} & \textbf{90.3} & \textbf{86.8} & \textbf{89.9} & \textbf{94.9} \\ \midrule
VNI-CT-FGSM & 94.1          & 92.4          & 91.6          & 95.1          & 92.2          & 92.3          & 84.5          & 77.4          & 81.4          & 89.0          \\
GI-VNI-CT-FGSM & \textbf{98.9} & \textbf{98.1} & \textbf{98.0} & \textbf{99.1} & \textbf{97.8} & \textbf{98.2} & \textbf{90.8} & \textbf{87.2} & \textbf{90.5} & \textbf{95.4} \\ \bottomrule
\end{tabular}}
\caption{Attack success rates (\%) of nine advanced defense methods using both optimization methods and transformation methods. The adversarial examples are crafted by the ensemble of Inc-v3, Inc-v4, IncRes-v2, Res-101 models.}
\label{tab4}
\end{table*}
\subsection{Attack with Video Models}
To verify the generality of our method, we further validate the experimental performance in video recognition attacks. Following the experimental setup of ~\citep{wei2022boosting}, we choose UCF-101~\citep{soomro2012ucf101} for the dataset, NL~\citep{wang2018non}, SlowFast~\citep{feichtenhofer2019slowfast} and TPN~\citep{yang2020temporal} for the video recognition models, and Resnet-50 and Resnet-101 are chosen as the backbone for each network, respectively.

We conduct experiments under TT~\citep{wei2022boosting}, a powerful baseline for transfer attacks against video recognition models, FGSM~\citep{fgsm}, and the proposed method (GI) in this section, respectively. Following the standard experimental setup of ~\citep{wei2022boosting}, we set the number of iterations $T = 1$ and the maximum perturbation $\epsilon = 16$. The experimental results are shown in Table~\ref{video}, which shows that the proposed method in this section is also extremely effective in attacking the video recognition model, and can improve the success rate of the transfer attack by more than 15\% on average. Overall, this section confirms the effectiveness of the proposed method in video recognition attacks, and also fully illustrates the generalization and effectiveness of the proposed method in attacks in different domains.

\begin{table*}[]
\renewcommand{\arraystretch}{0.82} 
\begin{center}
\setlength{\tabcolsep}{1.3mm}
\scalebox{0.7}{
\begin{tabular}{@{}c|c|ccccccc@{}}
\toprule
Model                            & Attack     & NL-Res101      & NL-Res50     & SlowFast-Res101 & SlowFast-Res50 & TPN-Res101     & TPN-Res50     & Avg.          \\ \midrule
\multirow{4}{*}{NL-Res101}      & FGSM       & 77.2*           & 84.2          & 37.6            & 46.5           & 13.9           & 21.8          & 46.9          \\
                                 & GI-FGSM    & \textbf{100.0*} & \textbf{91.3} & 34.6            & 49.3           & 15.8           & 18.8          & 51.6          \\
                                 & TT         & 72.3*           & 72.3          & 49.5            & 57.4           & 30.7           & 44.5          & 54.5          \\
                                 & GI-TT-FGSM & 99.0*           & 87.8          & \textbf{61.3}   & \textbf{69.3}  & \textbf{43.3}  & \textbf{57.9} & \textbf{69.7} \\ \midrule
\multirow{4}{*}{SlowFast-Res101} & FGSM       & 42.6            & 58.4          & 79.2*           & 63.4           & 28.7           & 38.6          & 51.8          \\
                                 & GI-FGSM    & 41.5            & 57.6         & \textbf{99.0*}  & \textbf{81.9}  & 29.7           & 39.6          & 58.2          \\
                                 & TT         & 58.4            & 65.3          & 71.3*           & 58.4           & 29.7           & 41.6          & 54.1          \\
                                 & GI-TT-FGSM & \textbf{77.4}   & \textbf{82.1} & 98.0*           & 81.1           & \textbf{40.5}  & \textbf{60.4} & \textbf{73.3} \\ \midrule
\multirow{4}{*}{TPN-Res101}      & FGSM       & 57.4            & 67.3          & 49.5            & 46.5           & 36.6*          & 47.5          & 50.8          \\
                                 & GI-FGSM    & \textbf{68.7}   & 76.4          & 45.5            & 50.8           & 88.1*          & 59.6          & 64.8          \\
                                 & TT         & 55.5            & 65.3          & 47.5            & 47.5           & 46.5*          & 52.5          & 52.5          \\
                                 & GI-TT-FGSM & 68.3            & \textbf{79.8} & \textbf{52.4}   & \textbf{53.4}  & \textbf{96.0*} & \textbf{67.3} & \textbf{69.5} \\ \bottomrule
\end{tabular}
}
\caption{\footnotesize Attack success rates (\%) under video models. The adversarial examples are crafted by NL-Res101, SlowFast-Res101, TPN-Res101 models respectively. * indicates the white box attack setting.}
\vspace{-16pt}
\label{video}
\end{center}
\end{table*}

\subsection{Ablation Study}
In this section, we conduct a series of ablation experiments on the hyperparameters. We consider Inc-v3 to generate adversarial examples under the CT method and transfer them to all models, and then do separate ablation experiments on pre-convergences as well as the search amplification factor. All experimental hyperparameters of the other methods are kept constant.
\begin{figure*}[t]
	\centering
	\subfloat[MI-CT-FGSM]{\label{fig:mipre}\includegraphics[width = 0.28\textwidth]{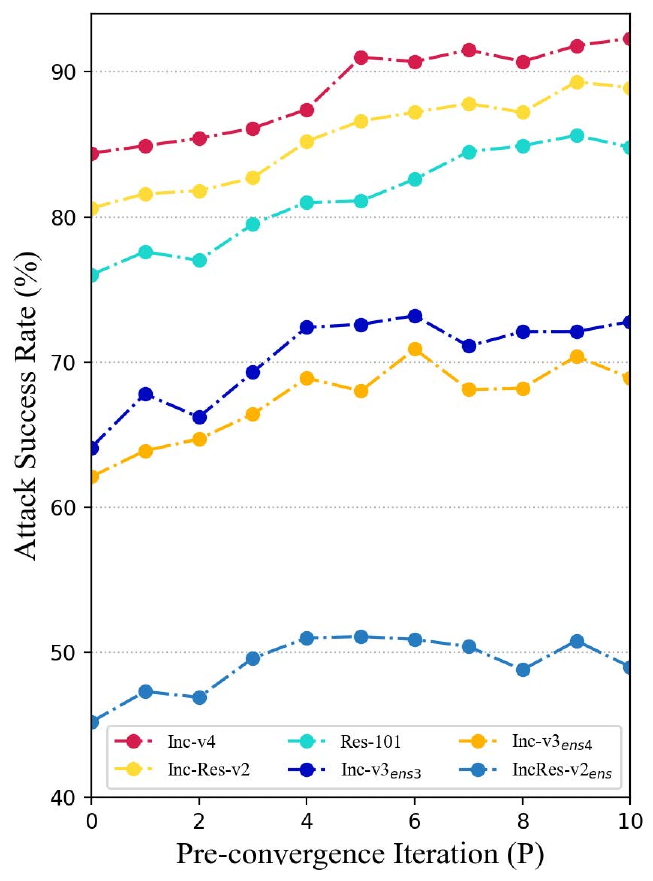}}\hfill 
	\subfloat[NI-CT-FGSM]{\label{fig:nipre}\includegraphics[width = 0.28\textwidth]{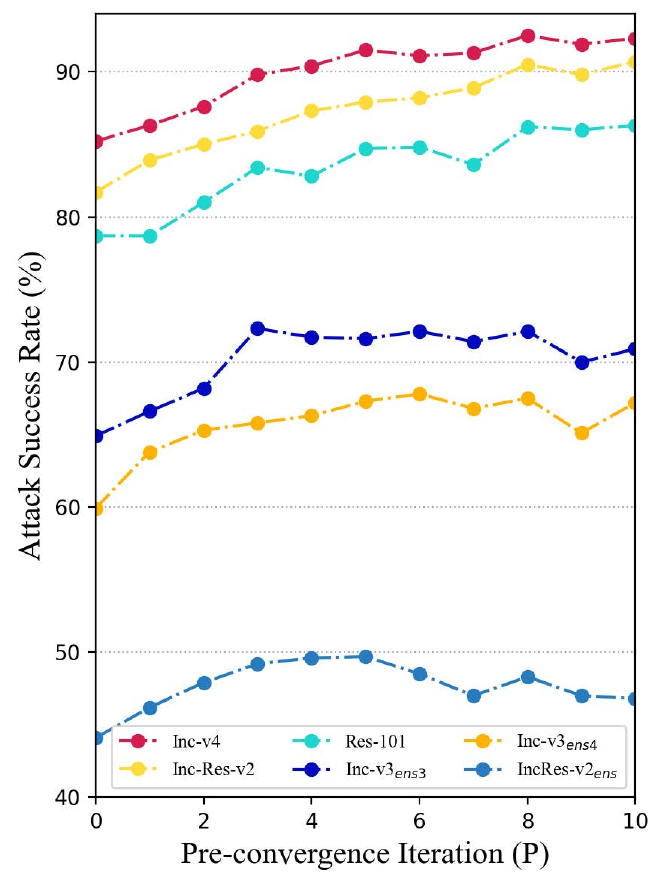}}\hfill
    \subfloat[]{\label{fig:amp2}\includegraphics[width = 0.33\textwidth]{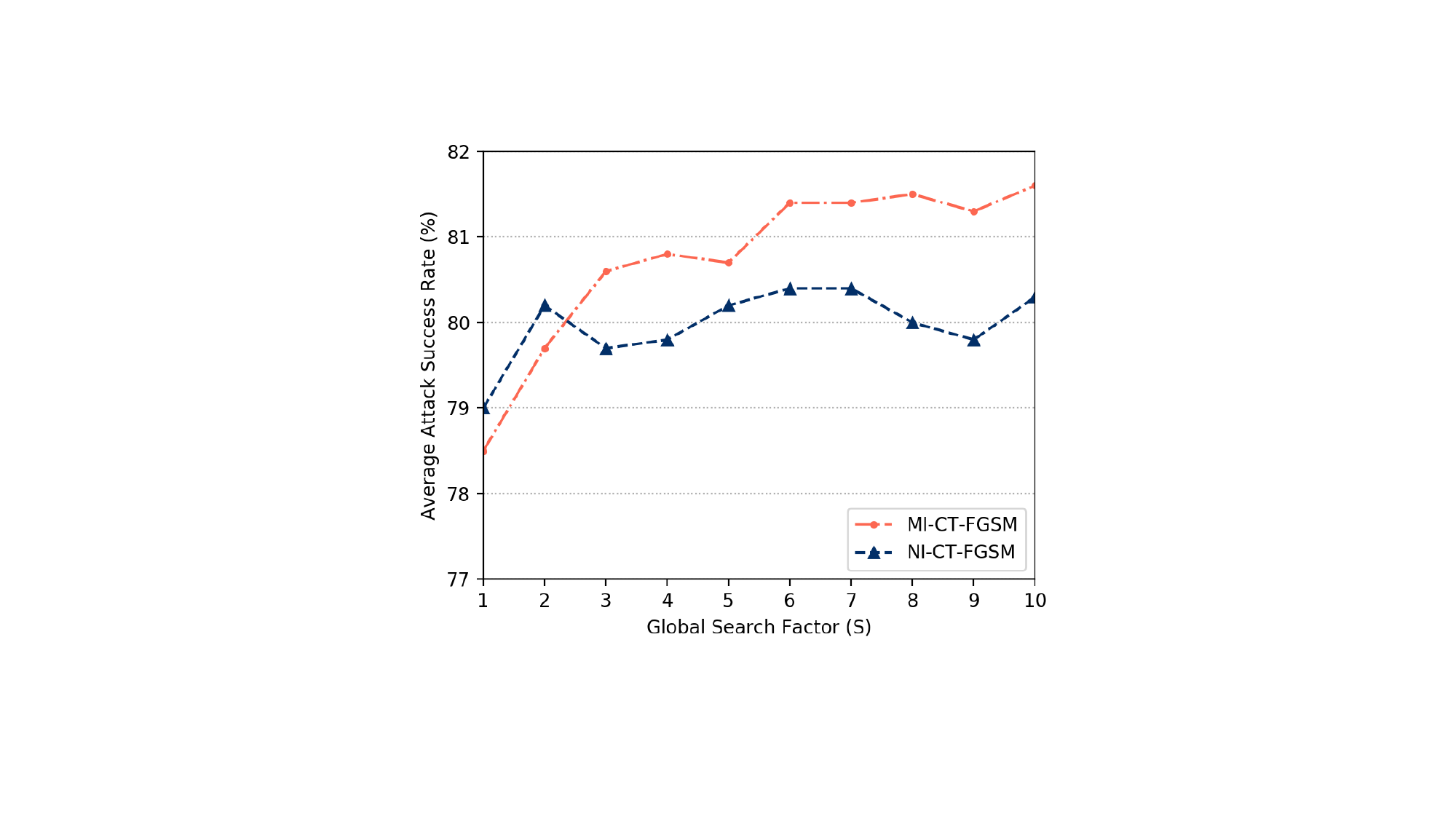}}\\
	\caption{Ablation experiments on pre-convergence and global search factors. (a) and (b) represent attack success rate for six black box models using different pre-convergence iterations. (c) represents the average attack success rate of six models with different global search factors.}
    \label{fig:abl}
\end{figure*}
\textbf{Pre-convergence Iterations $P$}. Given the sequential necessity of both pre-convergence and global search, we first explore the effect of pre-convergence rounds $P$ on the attack results. As the experimental results in Figure~\ref{fig:mipre},~\ref{fig:nipre} demonstrate, we can find that just one step of pre-convergence yields a certain performance improvement that increases gradually with incremental $P$ until it stabilizes. This is a strong indication that there is gradient elimination in the unpreconverged gradient, and that pre-convergence suppresses this phenomenon, thus improving attack performance. Finally, we choose $P$ = 5 to balance the success rate of the attack with the time taken by the attack.

\textbf{Global Search Factor $S$}. After determining the number of pre-convergence iterations, we further investigate the effect of the global search factor at $P$ = 5, the results are shown in Figure~\ref{fig:amp2}. It can be seen that the attack performance improves significantly with the increase of the amplification factor and gradually stabilizes, which illustrates that the global search helps the momentum to find a better global optimal solution. In the end, we choose $S$ = 10.

\begin{table*}[t]
\begin{center}
\scalebox{0.7}{
\begin{tabular}{@{}c|cccccccc@{}}
\toprule
Attack        & Inc-v3         & Inc-v4        & IncRes-v2     & Res-101       & Inc-v3$_{ens3}$    & Inc-v3$_{ens4}$    & IncRes-v2$_{ens}$  & Avg-       \\ \midrule
MI-CT-FGSM   & 99.3*          & 85.8          & 84.1          & 87.5          & 48.0          & 43.5          & 28.1          & 68.1          \\
GI-MI-CT-FGSM & \textbf{99.4*} & \textbf{93.4} & \textbf{91.8} & \textbf{87.6} & \textbf{74.9} & \textbf{68.8} & \textbf{53.5} & \textbf{81.3} \\ \bottomrule
\end{tabular}}
\end{center}
\vspace{-8pt}
\caption{Attack success rate (\%) of MI-CT-FGSM and the combination with our method. Here the MI-CT-FGSM employs the same $P + T$ attack iterations to guarantee the same attack time as the our GI-MI-CT-FGSM. Adversarial examples are generated by Inc-v3.}
\label{time}
\end{table*}

For the sake of ensuring a fair comparison, we evaluate our method against a baseline that employs a strategy of employing a large step size during the initial $P$ steps and subsequently transitioning to a smaller step size for the subsequent $T$ steps. The results are shown in Table~\ref{time}. It can be seen that without our momentum initialization, even with a guaranteed higher number of iteration rounds, the attack success rate can only reach 68.1\%, while corresponding to our method we can achieve an attack success rate of 81.3\%. The absence of performance improvement when simply adding extra iteration rounds further highlights the significance of momentum initialization. We posit that the direct attack approach fails due to two main reasons. Firstly, the usage of an excessively large step size results in overfitting of the attack, as elaborated in detail in Table~\ref{fig:scale}. Secondly, the issue of gradient elimination remains unresolved, trapping the attack in a locally optimal solution due to the unconverged direction of the attack.

\section{Conclusion}
In this paper, we present an analysis of the limitations of existing optimization methods from the viewpoint of gradient consistency. We identify the issue of gradient elimination, which may hinder the effectiveness of forward attack attempts. Additionally, we observe that traditional iterative attacks rely on smaller steps that restrict the attack within a limited data distribution, leading to a higher probability of being trapped in a local optimum. Our proposal is to address these issues by global momentum initialization in the form of high-step size pre-convergence, which provides high-quality momentum knowledge for the attack during the initial stage, suppressing gradient elimination and enhancing its performance significantly. Our approach can be combined with almost all gradient-based optimization methods. Empirical experiments demonstrate that our method achieves an average success rate of 95.4\% with the nine existing advanced defense methods, significantly exceeding the previous state-of-the-art success rate of 89.0\%. Further, our method also shows impressive performance in video attacks. We hope our method can serve as a new baseline to validate the effectiveness of existing defense methods in the future.

\noindent\textbf{Limitation.} Most of the current transfer attack methods suffer from a significant drop in performance when under cross-model structure, and our method does not address this issue well. On the other hand, the current proposed method has only been validated in classification tasks, and its performance should be explored under more tasks.

\noindent{\textbf{Future Work. }Although GI-FGSM has been shown to be very effective in improving transferability, we believe there are still several aspects that deserve further exploration in the future. From a task perspective, GI-FGSM is currently only applicable to classification tasks, and the performance of the method under other tasks can be explored. From the attack process of the algorithm, further optimization of the algorithm can be considered by adding the pre-convergence process to the iterative attack to reduce the time cost. From a theoretical point of view, gradient consistency can be utilized to explore the generation of generalized perturbations.

\section*{CRediT authorship contribution statement}
Jiafeng Wang: Conceptualization, Investigation, Experimentation, Visualization, Writing-Original Draft, Review \& Editing. Zhaoyu Chen: Supervision, Experimentation, Writing-Original Draft, Review \& Editing. Kaixun Jiang: Investigation, Experimentation, Review \& Editing. Dingkang Yang: Investigation, Review \& Editing. Lingyi Hong: Investigation, Review \& Editing. Pinxue Guo: ~Review \& Editing. Haijing Guo: Review \& Editing. Wenqiang Zhang: Supervision, Review \& Editing, Funding acquisition.
\section*{Declaration of competing interest}
The authors declare that they have no known competing financial
interests or personal relationships that could have appeared to influence
the work reported in this paper. 
\section*{Data availability}
Data will be made available on request. 
\section*{Acknowledgments}
This work was supported by National Natural Science Foundation of China (No.62072112), Scientific and Technological innovation action plan of Shanghai Science and Technology Committee (No.22511102202), Fudan Double First-class Construction Fund (No. XM03211178).

\bibliographystyle{model5-names}

\bibliography{eswa2024}

\end{document}